\pdfoutput=1

\documentclass[11pt]{article}

\usepackage{acl}

\usepackage{times}
\usepackage{latexsym}
\usepackage{graphicx}
\usepackage{tabularx}
\usepackage{siunitx}
\usepackage{algorithm}
\usepackage{algpseudocode}
\usepackage[T1]{fontenc}

\usepackage[utf8]{inputenc}

\usepackage{microtype}

\usepackage{inconsolata}
\usepackage{booktabs}
\usepackage{multirow}
\usepackage{examples-slim}

\newcommand{\eg}[1]{(\ref{#1})}

\usepackage[]{titlesec}
\titlespacing*{\paragraph}{\parindent}{0ex}{1ex}

\title{Detecting Contradictory COVID-19 Drug Efficacy Claims \\[0.5ex]
from Biomedical Literature\thanks{{}~Data and code available at \url{https://github.com/dnsosa/covid_lit_contra_claims}}}

\newcommand{\ourauthorspace}{\hspace{12pt}}

\author{Daniel N.~Sosa$^1$
\ourauthorspace
Malavika Suresh$^2$
\ourauthorspace Christopher Potts$^3$
\ourauthorspace
Russ B.~Altman$^{4,5}$\\[1ex]
        $^1$Department of Biomedical Data Science, Stanford University\\
        $^2$School of Computing, Robert Gordon University\\
        $^3$Department of Linguistics, Stanford University\\
        $^4$Department of Bioengineering, Stanford University\\
        $^5$Department of Genetics, Stanford University\\
        \texttt{\{dnsosa,cgpotts,russ.altman\}@stanford.edu}
        \qquad
        \texttt{m.suresh@rgu.ac.uk}
        }

\begin{document}
\maketitle
\begin{abstract}
The COVID-19 pandemic created a deluge of questionable and contradictory scientific claims about drug efficacy -- an ``infodemic'' with lasting consequences for science and society. In this work, we argue that NLP models can help domain experts distill and understand the literature in this complex, high-stakes area. Our task is to automatically identify contradictory claims about COVID-19 drug efficacy. We frame this as a natural language inference problem and offer a new NLI dataset created by domain experts. The NLI framing allows us to create curricula combining existing datasets and our own. The resulting models are useful investigative tools. We provide a case study of how these models help a domain expert summarize and assess evidence concerning remdisivir and hydroxychloroquine.
\end{abstract}

\section{Introduction}

The COVID-19 pandemic caused by the novel SARS-CoV-2 virus completely changed modern life. According to the World Health Organization Nov.~16, 2022 situation report, more than 6.5 million people have died as a result of this disease \cite{WHOreport}. During times of pandemic, treatment options are limited, and developing new drug treatments is infeasible in the short-term \cite{WoutersDrugDev}.

However, if a novel disease shares biological underpinnings with another disease for which a drug treatment already exists, a doctor may be able to repurpose that drug as a treatment for the new disease with positive therapeutic effect \cite{PushpakomRepurposingReview}. This strategy has been successful in several contexts \cite{repurposingReview2, repurposingExample2, repurposingExample3} and may be the only viable strategy during an emerging pandemic.

Decisions about repurposing drug treatments are predicated on scientific knowledge. Making predictions about how to repurpose an existing drug requires understanding the target disease's mechanism. Because SARS-CoV-2 was a new virus, our knowledge of COVID-19's mechanism rapidly evolved. The biomedical literature about the virus and disease proliferated at an unprecedented rate \cite{IoannidisCovidization, ioannidisCovidAuthors}. The need for knowledge about the virus and the bottleneck of limited peer reviewers led to many cases of circumventing typical quality control mechanisms for research. To inform their clinical practice, healthcare professionals relied on knowledge sources of lower scientific quality including early clinical reports with small sample sizes and non-peer reviewed manuscripts posted on preprint servers \cite{NouriCOVIDPreprints}. This deluge of rapidly changing information became an ``infodemic'', and it became infeasible for the average clinician to stay up-to-date with the growing literature \cite{LancetInfodemic}.

Automated methods have great potential to help domain experts fight such an infodemic. We illustrate this potential with a case study focused on automatically detecting contradictory research claims in the COVID-19 therapeutics literature. We frame this as a natural language inference (NLI) problem: given pairs of research claims in biomedical literature, we develop models that predict whether they entail, contradict, or are neutral with respect to each other. Our models are trained on a new dataset of these claim pairs extracted from the CORD-19 dataset \citep{cord19} and annotated by domain experts. Our best models are trained on curricula \citep{curriculum} of existing NLI datasets and our domain-specific one. These models are effective at the NLI task, but the ultimate test of their value is whether they can help domain experts. We show how these models can help a domain expert to see early on that hydroxychloroquine was an ineffective COVID-19 treatment and how the story of remdisivir was still emerging.

\section{COVID-19 NLI Dataset}\label{sec:dataset}

Our new COVID-19 NLI dataset consists of pairs of research claims describing COVID-19 drug treatment efficacy and safety. These claims came from the subset of the June 17, 2020 (v33) CORD-19 \citep{cord19} manuscripts containing a COVID-19-related term (e.g., ``SARS-CoV-2'', ``2019-nCov''). Claims were extracted from the articles' full text using the LSTM approach of \citet{achakulvisut}. False positive research claims were manually removed.

To begin the annotation process, we inspected pairs of claims on common drugs and topics. This led us to a set of five categories:
Strict Entailment,
Entailment,
Possible Entailment,
Strict Contradiction,
Contradiction, and
Neutral.
Our annotation guidelines were developed and refined by clinically trained annotators (nurses and a biomedical researcher) over two preliminary rounds of annotation. In Round~1, four annotators labeled 64 claim pairs (Fleiss'~$\kappa$ = 0.83). The team discussed this process and refined the guidelines. In Round~2, three annotators (a subset of those from Round~1) annotated 75 claim pairs (Fleiss'~$\kappa$ = 0.84) using the new guidelines, and then determined that they were ready to scale (Appendix~\ref{sec:iaaAndGuidelines}).

For the dataset itself, 1000 pairs of claims were sampled for annotation using three criteria: (1) both claims mention at least one of 7 treatment candidates (\{\text{``hydroxychloroquine''}, \text{``chloroquine''}, \text{``tocilizumab''}, \text{``remdesivir''}, \text{``vitamin D''}, \text{``lopinavir''}, \text{``dexamethasone''}\}), (2) high similarity between the claim's embedding and the embedding for a word in a predefined topic list (\{``mortality'', ''effective treatment'', ``toxicity''\}), using uSIF embeddings \citep{usif}, and (3) non-zero polarities of equal or opposite sign using VADER \citep{vader}. Appendix~\ref{sec:claimPrep} provides further details.

Each annotation requires a large time investment from the annotator and draws heavily on their domain expertise, so each example was annotated by a single annotator, with the inter-annotator agreement rounds and guidelines serving to ensure consistency across the dataset.

Because some claims are present in multiple claim pairs, we selected a subset of pairs such that no claim is present in more than one train, validation, or test split to prevent test-set leakage. From the network of claim pairs (claims are nodes, and co-occurrences in an annotated pair are edges), we selected 3 disjoint subnetworks to comprise the train, validation, and test splits. The resulting dataset contains 778 total claim pairs.

\section{Model Development}

Our goal is to develop a model to help domain experts find and adjudicate contradictory claims in the COVID-19 literature. We explored a wide range of techniques for developing the best model given our available data. The Appendices provide a full overview of all these experiments and comparisons. Here, we provide a high-level summary.

\paragraph{Pretrained Parameters}
All our models begin with pretrained parameters created using the general architecture of BERT \citep{bert}. Five pre-trained BERT models were evaluated for further fine-tuning: PubMedBERT \citep{pubmedbert}, SciBERT \citep{scibert}, BioBERT \citep{biobert}, BioClinBERT \citep{bioclinbert}, and RoBERTa \cite{roberta}. We found that PubMedBERT was the best for our task across all fine-tuning regimes (Appendix~\ref{sec:bertPretraining}).

\paragraph{Fine-tuning Curricula}
For fine-tuning these parameters, we use MultiNLI \citep{multinli}, MedNLI \citep{mednli}, ManConCorpus \citep{manConCorpus}, and our new COVID-19 NLI Dataset (with our six labels collapsed to three as in the other datasets). We found that the best models were achieved with a curriculum that arranged these in the order we gave above. This is intuitively an arrangement from most general to most domain specific, which aligns with existing results and intuitions for curriculum learning \citep{curriculum,nlucurriculum,pretraincurriculum}. For detailed descriptions of these datasets, the range of curricula we explored, and our procedures for hyperparameter tuning, we refer to Appendices~\ref{sec:curriculumDatasets}, \ref{sec:curriculumDesign}, and \ref{sec:hpOpt}, respectively.

\paragraph{Results}
To contextualize our results on this hard, novel task, we evaluated a number of baselines using sparse feature representations and simple similarity calculations, as well as hypothesis-only variants of these models and our BERT models. These baselines are described in Appendix~\ref{sec:testSetEval}.

\begin{table}[tp]
\resizebox{\columnwidth}{!}{
\setlength{\tabcolsep}{4pt}
\begin{tabular}{@{} l l r r @{}}
\toprule
 & & & Contra. \\
Model & Curriculum & F1 &  Recall \\
\midrule
\multirow{3}{*}{PubMedBERT} & Forward & \textbf{0.690} & \textbf{0.571} \\
& Reverse & 0.428 & 0.381 \\
& Shuffled & 0.523 & 0.416 \\
\midrule
\multirow{3}{*}{RoBERTa} & Forward & 0.544 & 0.429 \\
 & Reverse & 0.411 & 0.476 \\
 & Shuffled & 0.239 & 0.119 \\
\midrule
PubMedBERT Hyp.~only & Forward & 0.485 & 0.190 \\
RoBERTa Hyp.~only & Forward & 0.433 & 0.095 \\
\bottomrule
\end{tabular}
}
\caption{Core results. Figure~\ref{fig:testMetrics} and Table~\ref{tab:RevShuffBERTs} expand these results to include a number of other baselines, most of which perform near chance. Metrics for the shuffled category are averages of the 4 shuffled curricula.}
\label{tab:results}
\end{table}

Table~\ref{tab:results} summarizes our results. We report F1 scores as well as Contradictions Recall, an important category for our case study. The best performance is achieved by the PubMedBERT model trained with the forward curriculum where fine-tuning takes place from general domain to complex, in-domain datasets. This setting outperforms baselines and alternative curricula by a large margin.

\section{Case Study: Wading Through the Sea of Drug Treatment Literature}

The value of our model lies in its potential to help domain experts tackle an infodemic. Our biomedical research group used the model to understand the state of knowledge about the efficacy and mechanism of two controversial treatments, hydroxychlorouqine and remdesivir, from the perspective of June 2020.

We first extracted all claims identified from COVID-19 manuscripts concerning a drug treatment, using the same procedure as for our COVID NLI dataset (Section~\ref{sec:dataset}), and we filtered that set to pairs of claims that were (1) sufficiently similar (uSIF similarity > 0.5) and (2) both mentioned remdesivir or hydroxychloroquine. We sampled pairs from 50 papers yielding 5,336 total pairs. We then used our best model to make predictions about all these pairs resulting in 322 predicted contradictions. We ranked these by the model's predicted probability of this class, and we inspected the highest probability predictions.

For remdesivir, one claim of limited efficacy from an clinical trial of 233 participants yielded several predicted contradictions:
\begin{examples}
\item\label{remcontra}
Remdesivir did not result in significant reductions in SARS-CoV-2 RNA loads or detectability in upper respiratory tract or sputum specimens in this study despite showing strong antiviral effects in preclinical models of infection with coronaviruses \cite{remNegative}.
\end{examples}
Nineteen unique papers contained a claim that was predicted to contradict this claim -- already a striking pattern that might have taken a researcher days to discover by hand by probing full text articles.

The specific claims that contradict our core claim are illuminating. One reads,
\begin{examples}
\item
The present study reveals that remdesivir has the highest potential in binding and therefore competitively inhibiting RDRP of SARS-CoV-2, among all known RDRP inhibitors \cite{remRNAInhibit},
\end{examples}
indicating strong chemical and pharmacodynamic reasoning supporting a mechanism of action for remdesivir. A second claim describes:
\begin{examples}
\item
Remdesivir treatment in rhesus macaques infected with SARS-CoV-2 was highly effective in reducing clinical disease and damage to the lungs \cite{remMacaques},
\end{examples}
surfacing particularly strong pre-clinical evidence. From another ongoing clinical trial including 1,064 patients, authors note:
\begin{examples}
\item
Preliminary results of this trial suggest that a 10-day course of remdesivir was superior to placebo in the treatment of hospitalized patients with COVID-19.
 \cite{remPositive}
\end{examples}
Overall, we are quickly able to glean how evidence supporting the remdesivir hypothesis was strong from a variety of pre-clinical studies in vastly different settings in 2020. Our original negative claim \eg{remcontra} presents real evidence against the drug. Still, though, the clinical picture was not yet clear, suggesting the need for further clinical investigation or better striation of populations or therapeutic windows for seeing efficacy.

For hydroxychloroquine, one of the earliest drugs considered, a different picture emerges. We focus in on a claim from a medRxiv preprint \eg{hydrocontra}:
\begin{examples}
\item\label{hydrocontra}
In summary, this retrospective study demonstrates that hydroxychloroquine application is associated with a decreased risk of death in critically ill COVID-19 patients without obvious toxicity and its mechanisms of action is probably mediated through its inhibition of inflammatory cytokine storm on top of its ability in inhibiting viral replication. \cite{hcqPositive}
\end{examples}

From its predicted contradictions, we immediately identified two clinical studies:

\begin{examples}\setlength{\itemsep}{0pt}
\item
Overall, these data do not support the addition of hydroxychloroquine to the current standard of care in patients with persistent mild to moderate COVID-19 for eliminating the virus. \cite{hcqNeg1}
\item
Although a marginal possible benefit from prophylaxis in a more at-risk group cannot be ruled out, the potential risks that are associated with hydroxychloroquine may also be increased in more at-risk populations, and this may essentially negate any benefits that were not shown in this large trial involving younger, healthier participants. \cite{hcqNeg2}
\end{examples}

These claims reflect the challenging language typical for the domain including hedging, multiple clauses, important context qualifiers (subpopulations and adverse events), and positive and negative sentiments. From these surfaced contradictions, we find evidence of the drug's inefficacy in mild and moderate cases and are led to discover the early observations of cardiac arrest being associated with hydroxychloroquine treatment. Again, discovering these claims \textit{de novo} is difficult given the size of the corpus of COVID-19 literature. Our NLI model greatly speeds up the process and allows domain experts to home in directly on relevant evidence.

\section{Stakeholders}

There are several biomedical stakeholders who would benefit from models like ours.

\paragraph{Epidemiologists}
Epidemiologists survey public health data to inform  policy decisions in collaboration with authoritative bodies like the NIH and WHO. Their recommendations must be conservative, so surfacing results that dispute claims of drug efficacy is critical. Their gold standard resource for aggregating evidence is the meta-analysis, but in the early stages of the pandemic, large randomized controlled trials (RCTs) had not completed, and review articles quickly became outdated.

\paragraph{FDA Regulators} Regulators too need to make conservative recommendations, as FDA approval signals to clinicians that a treatment is standard-of-care. Surfacing contradictory claims of drug efficacy and safety is essential \cite{fdaCOVIDdrugs}.

\paragraph{Researchers} By identifying areas of scientific uncertainty via contradictory evidence at all stages of the pipeline (\textit{in silico}, \textit{in vitro}, \textit{in vivo}, clinical), researchers could have more quickly identified fruitful areas of investigation \cite{repurposingBMI}.

\paragraph{Drug Manufacturers}
Manufacturers of repurposing candidates were incentivized to understand in what settings their drug seemed to be effective and by what mechanism. For claims of inefficacy, they were interested in surfacing any mitigating factors qualifying these claims or motivating follow-up analyses.

\section{Discussion and Conclusion}

In settings where the scale of literature is insurmountable for human readers, as is the case during a pandemic, automated curatorial assistants can be transformative \cite{coronaCentral}. During COVID-19, meta-analyses and review articles, which are written to synthesize a large body of literature, could not be comprehensive or quickly became outdated. In some cases, it was necessary to create meta-meta-analyses involving hundreds of papers \cite{covidMetametaanalysis}.

Our work shows the value of integrating NLP into the domain of meta-science, embracing all the complexities of biomedical research as it naturally exists in literature.  We presented an NLI framing for identifying contradictory or corroborating research claims in the challenging domain of COVID-19 drug efficacy. We created a new dataset and designed curricula for optimizing language model fine-tuning for the task. To illustrate the potential of our model, we showed that we were quickly able to distill the state of knowledge about hydroxychlorouqine and remdesivir efficacy as of June 2020, arriving at conclusions that are extremely well-supported in 2022.

Identifying where science is not consistent is necessary for understanding the current state of human knowledge and reveals frontiers for further research. Significant contradictions can often be found buried in biomedical articles; surfacing these instances nearly as quickly as research is publicly disseminated can generate leads that researchers and curators should pursue. Beyond facilitating search and discovery, our method can help estimate confidence in the consensus of facts in science when creating general knowledge representations \cite{contextsContras} for downstream applications like predicting novel drug repurposing opportunities \textit{in silico} \cite{litKGrepurposing}.

\section*{Acknowledgements}

Special thanks to our clinical annotation team for their invaluable contributions to this work. DS is supported by NIH training grant LM007033, Stanford Data Science, and BenevolentAI. RBA is supported by NIGMS GM102365, HG010615, and Chan Zuckerberg Biohub.

\bibliography{anthology,custom}

\begin{thebibliography}{43}
\expandafter\ifx\csname natexlab\endcsname\relax\def\natexlab#1{#1}\fi

\bibitem[{Achakulvisut et~al.(2020)Achakulvisut, Bhagavatula, Acuna, and
  Kording}]{achakulvisut}
Titipat Achakulvisut, Chandra Bhagavatula, Daniel Acuna, and Konrad Kording.
  2020.
\newblock \href {https://arxiv.org/abs/1907.00962} {Claim extraction in
  biomedical publications using deep discourse model and transfer learning}.
\newblock \emph{arXiv:1907.00962}.

\bibitem[{Al-Saleem et~al.(2021)Al-Saleem, Granet, Ramakrishnan, Ciancetta,
  Saveson, Gessner, and Zhou}]{repurposingExample3}
Jacob Al-Saleem, Roger Granet, Srinivasan Ramakrishnan, Natalie~A. Ciancetta,
  Catherine Saveson, Chris Gessner, and Qiongqiong Zhou. 2021.
\newblock \href {https://doi.org/10.1021/acs.jcim.1c00642} {Knowledge
  graph-based approaches to drug repurposing for {COVID}-19}.
\newblock \emph{Journal of Chemical Information and Modeling},
  61(8):4058--4067.

\bibitem[{Alamri and Stevenson(2016)}]{manConCorpus}
Abdulaziz Alamri and Mark Stevenson. 2016.
\newblock \href {https://doi.org/10.1186/s13326-016-0083-z} {A corpus of
  potentially contradictory research claims from cardiovascular research
  abstracts}.
\newblock \emph{Journal of Biomedical Semantics}, 7.

\bibitem[{Alsentzer et~al.(2019)Alsentzer, Murphy, Boag, Weng, Jindi, Naumann,
  and McDermott}]{bioclinbert}
Emily Alsentzer, John Murphy, William Boag, Wei-Hung Weng, Di~Jindi, Tristan
  Naumann, and Matthew McDermott. 2019.
\newblock \href {https://doi.org/10.18653/v1/W19-1909} {Publicly available
  clinical {BERT} embeddings}.
\newblock In \emph{Proceedings of the 2nd Clinical Natural Language Processing
  Workshop}, pages 72--78, Minneapolis, Minnesota, USA. Association for
  Computational Linguistics.

\bibitem[{Beigel et~al.(2020)Beigel, Tomashek, Dodd, Mehta, Zingman, Kalil,
  Hohmann, Chu, Luetkemeyer, Kline, Lopez~de Castilla, Finberg, Dierberg,
  Tapson, Hsieh, Patterson, Paredes, Sweeney, Short, Touloumi, Lye, Ohmagari,
  Oh, Ruiz-Palacios, Benfield, Fätkenheuer, Kortepeter, Atmar, Creech,
  Lundgren, Babiker, Pett, Neaton, Burgess, Bonnett, Green, Makowski, Osinusi,
  Nayak, and Lane}]{remPositive}
John~H. Beigel, Kay~M. Tomashek, Lori~E. Dodd, Aneesh~K. Mehta, Barry~S.
  Zingman, Andre~C. Kalil, Elizabeth Hohmann, Helen~Y. Chu, Annie Luetkemeyer,
  Susan Kline, Diego Lopez~de Castilla, Robert~W. Finberg, Kerry Dierberg,
  Victor Tapson, Lanny Hsieh, Thomas~F. Patterson, Roger Paredes, Daniel~A.
  Sweeney, William~R. Short, Giota Touloumi, David~Chien Lye, Norio Ohmagari,
  Myoung-don Oh, Guillermo~M. Ruiz-Palacios, Thomas Benfield, Gerd
  Fätkenheuer, Mark~G. Kortepeter, Robert~L. Atmar, C.~Buddy Creech, Jens
  Lundgren, Abdel~G. Babiker, Sarah Pett, James~D. Neaton, Timothy~H. Burgess,
  Tyler Bonnett, Michelle Green, Mat Makowski, Anu Osinusi, Seema Nayak, and
  H.~Clifford Lane. 2020.
\newblock \href {https://doi.org/10.1056/NEJMoa2007764} {Remdesivir for the
  treatment of {COVID}-19 a final report}.
\newblock \emph{New England Journal of Medicine}, 383(19):1813--1826.

\bibitem[{Beltagy et~al.(2019)Beltagy, Lo, and Cohan}]{scibert}
Iz~Beltagy, Kyle Lo, and Arman Cohan. 2019.
\newblock \href {https://doi.org/10.18653/v1/D19-1371} {{SciBERT}: A pretrained
  language model for scientific text}.
\newblock In \emph{Proceedings of the 2019 Conference on Empirical Methods in
  Natural Language Processing and the 9th International Joint Conference on
  Natural Language Processing ({EMNLP}-{IJCNLP})}, pages 3615--3620, Hong Kong,
  China. Association for Computational Linguistics.

\bibitem[{Bengio et~al.(2009)Bengio, Louradour, Collobert, and
  Weston}]{curriculum}
Yoshua Bengio, Jérôme Louradour, Ronan Collobert, and Jason Weston. 2009.
\newblock \href {https://doi.org/10.1145/1553374.1553380} {Curriculum
  learning}.
\newblock In \emph{Proceedings of the 26th Annual International Conference on
  Machine Learning}, {ICML} '09, pages 41--48, New York, NY, USA. Association
  for Computing Machinery.

\bibitem[{Borchers(2019)}]{fse}
Oliver Borchers. 2019.
\newblock Fast sentence embeddings.
\newblock \url{https://github.com/oborchers/Fast_Sentence_Embeddings}.

\bibitem[{Boulware et~al.(2020)Boulware, Pullen, Bangdiwala, Pastick, Lofgren,
  Okafor, Skipper, Nascene, Nicol, Abassi, Engen, Cheng, LaBar, Lother,
  MacKenzie, Drobot, Marten, Zarychanski, Kelly, Schwartz, McDonald,
  Rajasingham, Lee, and Hullsiek}]{hcqNeg2}
David~R. Boulware, Matthew~F. Pullen, Ananta~S. Bangdiwala, Katelyn~A. Pastick,
  Sarah~M. Lofgren, Elizabeth~C. Okafor, Caleb~P. Skipper, Alanna~A. Nascene,
  Melanie~R. Nicol, Mahsa Abassi, Nicole~W. Engen, Matthew~P. Cheng, Derek
  LaBar, Sylvain~A. Lother, Lauren~J. MacKenzie, Glen Drobot, Nicole Marten,
  Ryan Zarychanski, Lauren~E. Kelly, Ilan~S. Schwartz, Emily~G. McDonald, Radha
  Rajasingham, Todd~C. Lee, and Kathy~H. Hullsiek. 2020.
\newblock \href {https://doi.org/10.1056/NEJMoa2016638} {A randomized trial of
  hydroxychloroquine as postexposure prophylaxis for {COVID}-19}.
\newblock \emph{New England Journal of Medicine}, 383(6):517--525.

\bibitem[{Cassidy et~al.(2020)Cassidy, Dever, Stanbery, Edelman, Dworkin, and
  Nemunaitis}]{fdaCOVIDdrugs}
Christine Cassidy, Danielle Dever, Laura Stanbery, Gerald Edelman, Lance
  Dworkin, and John Nemunaitis. 2020.
\newblock \href {https://doi.org/10.1186/s13027-020-00338-z} {{FDA} efficiency
  for approval process of {COVID}-19 therapeutics}.
\newblock \emph{Infectious Agents and Cancer}, 15(1):73.

\bibitem[{Chivese et~al.(2021)Chivese, Musa, Hindy, Al-Wattary, Badran,
  Soliman, Aboughalia, Matizanadzo, Emara, Thalib, and
  Doi}]{covidMetametaanalysis}
Tawanda Chivese, Omran~A.H. Musa, George Hindy, Noor Al-Wattary, Saif Badran,
  Nada Soliman, Ahmed~T.M. Aboughalia, Joshua~T. Matizanadzo, Mohamed~M. Emara,
  Lukman Thalib, and Suhail~A.R. Doi. 2021.
\newblock \href {https://doi.org/10.1016/j.tmaid.2021.102135} {Efficacy of
  chloroquine and hydroxychloroquine in treating {COVID}-19 infection: A
  meta-review of systematic reviews and an updated meta-analysis}.
\newblock \emph{Travel Medicine and Infectious Disease}, 43:102135.

\bibitem[{Choudhury et~al.(2021)Choudhury, Moulick, Saikia, and
  Mazumder}]{remRNAInhibit}
Shuvasish Choudhury, Debojyoti Moulick, Purbajyoti Saikia, and
  Muhammed~Khairujjaman Mazumder. 2021.
\newblock \href {https://doi.org/10.1016/j.mjafi.2020.05.005} {Evaluating the
  potential of different inhibitors on {RNA}-dependent {RNA} polymerase of
  severe acute respiratory syndrome coronavirus 2: A molecular modeling
  approach}.
\newblock \emph{Medical Journal Armed Forces India}, 77:S373--S378.

\bibitem[{Corsello et~al.(2017)Corsello, Bittker, Liu, Gould, McCarren,
  Hirschman, Johnston, Vrcic, Wong, Khan, Asiedu, Narayan, Mader, Subramanian,
  and Golub}]{repurposingReview2}
Steven~M. Corsello, Joshua~A. Bittker, Zihan Liu, Joshua Gould, Patrick
  McCarren, Jodi~E. Hirschman, Stephen~E. Johnston, Anita Vrcic, Bang Wong,
  Mariya Khan, Jacob Asiedu, Rajiv Narayan, Christopher~C. Mader, Aravind
  Subramanian, and Todd~R. Golub. 2017.
\newblock \href {https://doi.org/10.1038/nm.4306} {The drug repurposing hub: a
  next-generation drug library and information resource}.
\newblock \emph{Nature Medicine}, 23(4):405--408.

\bibitem[{Devlin et~al.(2019)Devlin, Chang, Lee, and Toutanova}]{bert}
Jacob Devlin, Ming-Wei Chang, Kenton Lee, and Kristina Toutanova. 2019.
\newblock \href {https://doi.org/10.18653/v1/N19-1423} {{BERT}: Pre-training of
  deep bidirectional transformers for language understanding}.
\newblock In \emph{Proceedings of the 2019 Conference of the North {A}merican
  Chapter of the Association for Computational Linguistics: Human Language
  Technologies, Volume 1 (Long and Short Papers)}, pages 4171--4186,
  Minneapolis, Minnesota. Association for Computational Linguistics.

\bibitem[{Ethayarajh(2018)}]{usif}
Kawin Ethayarajh. 2018.
\newblock \href {https://doi.org/10.18653/v1/W18-3012} {Unsupervised random
  walk sentence embeddings: A strong but simple baseline}.
\newblock In \emph{Proceedings of the Third Workshop on Representation Learning
  for {NLP}}, pages 91--100, Melbourne, Australia. Association for
  Computational Linguistics.

\bibitem[{Gu et~al.(2021)Gu, Tinn, Cheng, Lucas, Usuyama, Liu, Naumann, Gao,
  and Poon}]{pubmedbert}
Yu~Gu, Robert Tinn, Hao Cheng, Michael Lucas, Naoto Usuyama, Xiaodong Liu,
  Tristan Naumann, Jianfeng Gao, and Hoifung Poon. 2021.
\newblock \href {https://doi.org/10.1145/3458754} {Domain-specific language
  model pretraining for biomedical natural language processing}.
\newblock \emph{ACM Transactions on Computing for Healthcare}, 3(1):2:1--2:23.

\bibitem[{Himmelstein et~al.(2022)Himmelstein, Lizee, Hessler, Brueggeman,
  Chen, Hadley, Green, Khankhanian, and Baranzini}]{repurposingExample2}
Daniel~Scott Himmelstein, Antoine Lizee, Christine Hessler, Leo Brueggeman,
  Sabrina~L Chen, Dexter Hadley, Ari Green, Pouya Khankhanian, and Sergio~E
  Baranzini. 2022.
\newblock \href {https://doi.org/10.7554/eLife.26726} {Systematic integration
  of biomedical knowledge prioritizes drugs for repurposing}.
\newblock \emph{eLife}, 6:e26726.

\bibitem[{Hutto and Gilbert(2014)}]{vader}
CJ~Hutto and Eric Gilbert. 2014.
\newblock \href {https://doi.org/10.1609/icwsm.v8i1.14550} {{VADER}: A
  parsimonious rule-based model for sentiment analysis of social media text}.
\newblock \emph{Proceedings of the International AAAI Conference on Web and
  Social Media}, 8(1):216--225.

\bibitem[{Ioannidis et~al.(2022{\natexlab{a}})Ioannidis, Bendavid,
  Salholz-Hillel, Boyack, and Baas}]{IoannidisCovidization}
John P.~A. Ioannidis, Eran Bendavid, Maia Salholz-Hillel, Kevin~W. Boyack, and
  Jeroen Baas. 2022{\natexlab{a}}.
\newblock \href {https://doi.org/10.1073/pnas.2204074119} {Massive covidization
  of research citations and the citation elite}.
\newblock \emph{Proceedings of the National Academy of Sciences},
  119(28):e2204074119.

\bibitem[{Ioannidis et~al.(2022{\natexlab{b}})Ioannidis, Salholz-Hillel,
  Boyack, and Baas}]{ioannidisCovidAuthors}
John P.~A. Ioannidis, Maia Salholz-Hillel, Kevin~W. Boyack, and Jeroen Baas.
  2022{\natexlab{b}}.
\newblock \href {https://doi.org/10.1098/rsos.210389} {The rapid, massive
  growth of {COVID}-19 authors in the scientific literature}.
\newblock \emph{Royal Society Open Science}, 8(9):210389.

\bibitem[{Lee et~al.(2020)Lee, Yoon, Kim, Kim, Kim, So, and Kang}]{biobert}
Jinhyuk Lee, Wonjin Yoon, Sungdong Kim, Donghyeon Kim, Sunkyu Kim, Chan~Ho So,
  and Jaewoo Kang. 2020.
\newblock \href {https://doi.org/10.1093/bioinformatics/btz682} {{BioBERT}: A
  pre-trained biomedical language representation model for biomedical text
  mining}.
\newblock \emph{Bioinformatics}, 36(4):1234--1240.

\bibitem[{Lever and Altman(2021)}]{coronaCentral}
Jake Lever and Russ~B. Altman. 2021.
\newblock \href {https://doi.org/10.1073/pnas.2100766118} {Analyzing the vast
  coronavirus literature with {CoronaCentral}}.
\newblock \emph{Proceedings of the National Academy of Sciences},
  118(23):e2100766118.

\bibitem[{Li et~al.(2013)Li, Zhou, Pei, Zhou, Li, Wu, and
  Hui}]{LiFishConsumption}
Yue-hua Li, Cheng-hui Zhou, Han-jun Pei, Xian-liang Zhou, Li-huan Li, Yong-jian
  Wu, and Ru-tai Hui. 2013.
\newblock \href
  {https://journals.lww.com/cmj/Fulltext/2013/03050/Fish_consumption_and_incidence_of_heart_failure__a.27.aspx}
  {Fish consumption and incidence of heart failure: A meta-analysis of
  prospective cohort studies}.
\newblock \emph{Chinese Medical Journal}, 126(5):942--948.

\bibitem[{Liu et~al.(2019)Liu, Ott, Goyal, Du, Joshi, Chen, Levy, Lewis,
  Zettlemoyer, and Stoyanov}]{roberta}
Yinhan Liu, Myle Ott, Naman Goyal, Jingfei Du, Mandar Joshi, Danqi Chen, Omer
  Levy, Mike Lewis, Luke Zettlemoyer, and Veselin Stoyanov. 2019.
\newblock \href {https://doi.org/10.48550/ARXIV.1907.11692} {{RoBERTa}: A
  robustly optimized {BERT} pretraining approach}.
\newblock \emph{arXiv:1907.11692}.

\bibitem[{Nagatsuka et~al.(2021)Nagatsuka, Broni-Bediako, and
  Atsumi}]{pretraincurriculum}
Koichi Nagatsuka, Clifford Broni-Bediako, and Masayasu Atsumi. 2021.
\newblock \href {https://aclanthology.org/2021.ranlp-1.112} {Pre-training a
  {BERT} with curriculum learning by increasing block-size of input text}.
\newblock In \emph{Proceedings of the International Conference on Recent
  Advances in Natural Language Processing ({RANLP} 2021)}, pages 989--996.

\bibitem[{Nouri et~al.(2021)Nouri, Cohen, Madhavan, Slomka, Iskandrian, and
  Einstein}]{NouriCOVIDPreprints}
Shayan~N. Nouri, Yosef~A. Cohen, Mahesh~V. Madhavan, Piotr~J. Slomka, Ami~E.
  Iskandrian, and Andrew~J. Einstein. 2021.
\newblock \href {https://doi.org/10.1111/jep.13498} {Preprint manuscripts and
  servers in the era of coronavirus disease 2019}.
\newblock \emph{Journal of Evaluation in Clinical Practice}, 27(1):16--21.

\bibitem[{Pushpakom et~al.(2019)Pushpakom, Iorio, Eyers, Escott, Hopper, Wells,
  Doig, Guilliams, Latimer, McNamee, Norris, Sanseau, Cavalla, and
  Pirmohamed}]{PushpakomRepurposingReview}
Sudeep Pushpakom, Francesco Iorio, Patrick~A. Eyers, K.~Jane Escott, Shirley
  Hopper, Andrew Wells, Andrew Doig, Tim Guilliams, Joanna Latimer, Christine
  McNamee, Alan Norris, Philippe Sanseau, David Cavalla, and Munir Pirmohamed.
  2019.
\newblock \href {https://doi.org/10.1038/nrd.2018.168} {Drug repurposing:
  Progress, challenges and recommendations}.
\newblock \emph{Nature Reviews Drug Discovery}, 18(1):41--58.

\bibitem[{Romanov and Shivade(2018)}]{mednli}
Alexey Romanov and Chaitanya Shivade. 2018.
\newblock \href {https://doi.org/10.18653/v1/D18-1187} {Lessons from natural
  language inference in the clinical domain}.
\newblock In \emph{Proceedings of the 2018 {Conference} on {Empirical}
  {Methods} in {Natural} {Language} {Processing}}, pages 1586--1596, Brussels,
  Belgium. Association for Computational Linguistics.

\bibitem[{Schardt et~al.(2007)Schardt, Adams, Owens, Keitz, and
  Fontelo}]{picoquestion}
Connie Schardt, Martha~B. Adams, Thomas Owens, Sheri Keitz, and Paul Fontelo.
  2007.
\newblock \href {https://doi.org/10.1186/1472-6947-7-16} {Utilization of the
  {PICO} framework to improve searching {PubMed} for clinical questions}.
\newblock \emph{BMC Medical Informatics and Decision Making}, 7(1):16.

\bibitem[{Sosa and Altman(2022)}]{contextsContras}
Daniel~N Sosa and Russ~B Altman. 2022.
\newblock \href {https://doi.org/10.1093/bib/bbac268} {Contexts and
  contradictions: A roadmap for computational drug repurposing with knowledge
  inference}.
\newblock \emph{Briefings in Bioinformatics}, 23(4):bbac268.

\bibitem[{Sosa et~al.(2021)Sosa, Chen, Kaushal, Lavertu, Lever, Rensi, and
  Altman}]{repurposingBMI}
Daniel~N. Sosa, Binbin Chen, Amit Kaushal, Adam Lavertu, Jake Lever, Stefano
  Rensi, and Russ Altman. 2021.
\newblock \href {https://doi.org/10.1016/j.jbi.2021.103673} {Repurposing
  biomedical informaticians for {COVID}-19}.
\newblock \emph{Journal of Biomedical Informatics}, 115:103673.

\bibitem[{Sosa et~al.(2020)Sosa, Derry, Guo, Wei, Brinton, and
  Altman}]{litKGrepurposing}
Daniel~N. Sosa, Alexander Derry, Margaret Guo, Eric Wei, Connor Brinton, and
  Russ~B. Altman. 2020.
\newblock \href {https://www.ncbi.nlm.nih.gov/pmc/articles/PMC6937428/} {A
  literature-based knowledge graph embedding method for identifying drug
  repurposing opportunities in rare diseases}.
\newblock \emph{Pacific Symposium on Biocomputing}, 25:463--474.

\bibitem[{Tang et~al.(2020)Tang, Cao, Han, Wang, Chen, Sun, Wu, Xiao, Liu,
  Chen, Chen, Wang, Yang, Lin, Zhao, Yan, Xie, Li, Yang, Liu, Qu, Ning, Shi,
  and Xie}]{hcqNeg1}
Wei Tang, Zhujun Cao, Mingfeng Han, Zhengyan Wang, Junwen Chen, Wenjin Sun,
  Yaojie Wu, Wei Xiao, Shengyong Liu, Erzhen Chen, Wei Chen, Xiongbiao Wang,
  Jiuyong Yang, Jun Lin, Qingxia Zhao, Youqin Yan, Zhibin Xie, Dan Li, Yaofeng
  Yang, Leshan Liu, Jieming Qu, Guang Ning, Guochao Shi, and Qing Xie. 2020.
\newblock \href {https://doi.org/10.1136/bmj.m1849} {Hydroxychloroquine in
  patients with mainly mild to moderate coronavirus disease 2019: open label,
  randomised controlled trial}.
\newblock \emph{BMJ}, 369:m1849.

\bibitem[{{The Lancet Infectious Diseases}(2020)}]{LancetInfodemic}
{The Lancet Infectious Diseases}. 2020.
\newblock \href {https://doi.org/10.1016/S1473-3099(20)30565-X} {The {COVID}-19
  infodemic}.
\newblock \emph{The Lancet Infectious Diseases}, 20(8):875.

\bibitem[{Wang et~al.(2020{\natexlab{a}})Wang, Lo, Chandrasekhar, Reas, Yang,
  Burdick, Eide, Funk, Katsis, Kinney, Li, Liu, Merrill, Mooney, Murdick,
  Rishi, Sheehan, Shen, Stilson, Wade, Wang, Wang, Wilhelm, Xie, Raymond, Weld,
  Etzioni, and Kohlmeier}]{cord19}
Lucy~Lu Wang, Kyle Lo, Yoganand Chandrasekhar, Russell Reas, Jiangjiang Yang,
  Doug Burdick, Darrin Eide, Kathryn Funk, Yannis Katsis, Rodney~Michael
  Kinney, Yunyao Li, Ziyang Liu, William Merrill, Paul Mooney, Dewey~A.
  Murdick, Devvret Rishi, Jerry Sheehan, Zhihong Shen, Brandon Stilson, Alex~D.
  Wade, Kuansan Wang, Nancy Xin~Ru Wang, Christopher Wilhelm, Boya Xie,
  Douglas~M. Raymond, Daniel~S. Weld, Oren Etzioni, and Sebastian Kohlmeier.
  2020{\natexlab{a}}.
\newblock \href {https://www.aclweb.org/anthology/2020.nlpcovid19-acl.1}
  {{CORD-19}: The {COVID-19} open research dataset}.
\newblock In \emph{Proceedings of the 1st Workshop on {NLP} for {COVID-19} at
  {ACL} 2020}. Association for Computational Linguistics.

\bibitem[{Wang et~al.(2020{\natexlab{b}})Wang, Zhang, Du, Du, Zhao, Jin, Fu,
  Gao, Cheng, Lu, Hu, Luo, Wang, Lu, Li, Wang, Ruan, Yang, Mei, Wang, Ding, Wu,
  Tang, Ye, Ye, Liu, Yang, Yin, Wang, Fan, Zhou, Liu, Gu, Xu, Shang, Zhang,
  Cao, Guo, Wan, Qin, Jiang, Jaki, Hayden, Horby, Cao, and Wang}]{remNegative}
Yeming Wang, Dingyu Zhang, Guanhua Du, Ronghui Du, Jianping Zhao, Yang Jin,
  Shouzhi Fu, Ling Gao, Zhenshun Cheng, Qiaofa Lu, Yi~Hu, Guangwei Luo,
  Ke~Wang, Yang Lu, Huadong Li, Shuzhen Wang, Shunan Ruan, Chengqing Yang,
  Chunlin Mei, Yi~Wang, Dan Ding, Feng Wu, Xin Tang, Xianzhi Ye, Yingchun Ye,
  Bing Liu, Jie Yang, Wen Yin, Aili Wang, Guohui Fan, Fei Zhou, Zhibo Liu,
  Xiaoying Gu, Jiuyang Xu, Lianhan Shang, Yi~Zhang, Lianjun Cao, Tingting Guo,
  Yan Wan, Hong Qin, Yushen Jiang, Thomas Jaki, Frederick~G. Hayden, Peter~W.
  Horby, Bin Cao, and Chen Wang. 2020{\natexlab{b}}.
\newblock \href {https://doi.org/10.1016/S0140-6736(20)31022-9} {Remdesivir in
  adults with severe {COVID}-19: A randomised, double-blind,
  placebo-controlled, multicentre trial}.
\newblock \emph{Lancet (London, England)}, 395(10236):1569--1578.

\bibitem[{Williams et~al.(2018)Williams, Nangia, and Bowman}]{multinli}
Adina Williams, Nikita Nangia, and Samuel Bowman. 2018.
\newblock \href {https://doi.org/10.18653/v1/N18-1101} {A broad-coverage
  challenge corpus for sentence understanding through inference}.
\newblock In \emph{Proceedings of the 2018 Conference of the North American
  Chapter of the Association for Computational Linguistics: Human Language
  Technologies, Volume 1 (Long Papers)}, pages 1112--1122, New Orleans,
  Louisiana. Association for Computational Linguistics.

\bibitem[{Williamson et~al.(2020)Williamson, Feldmann, Schwarz, Meade-White,
  Porter, Schulz, van Doremalen, Leighton, Yinda, Pérez-Pérez, Okumura,
  Lovaglio, Hanley, Saturday, Bosio, Anzick, Barbian, Cihlar, Martens, Scott,
  Munster, and de~Wit}]{remMacaques}
Brandi~N. Williamson, Friederike Feldmann, Benjamin Schwarz, Kimberly
  Meade-White, Danielle~P. Porter, Jonathan Schulz, Neeltje van Doremalen, Ian
  Leighton, Claude~Kwe Yinda, Lizzette Pérez-Pérez, Atsushi Okumura, Jamie
  Lovaglio, Patrick~W. Hanley, Greg Saturday, Catharine~M. Bosio, Sarah Anzick,
  Kent Barbian, Tomas Cihlar, Craig Martens, Dana~P. Scott, Vincent~J. Munster,
  and Emmie de~Wit. 2020.
\newblock \href {https://doi.org/10.1038/s41586-020-2423-5} {Clinical benefit
  of remdesivir in rhesus macaques infected with {SARS}-{C}o{V}-2}.
\newblock \emph{Nature}, 585(7824):273--276.

\bibitem[{Wolf et~al.(2020)Wolf, Debut, Sanh, Chaumond, Delangue, Moi, Cistac,
  Rault, Louf, Funtowicz, Davison, Shleifer, von Platen, Ma, Jernite, Plu, Xu,
  Le~Scao, Gugger, Drame, Lhoest, and Rush}]{huggingface}
Thomas Wolf, Lysandre Debut, Victor Sanh, Julien Chaumond, Clement Delangue,
  Anthony Moi, Pierric Cistac, Tim Rault, Remi Louf, Morgan Funtowicz, Joe
  Davison, Sam Shleifer, Patrick von Platen, Clara Ma, Yacine Jernite, Julien
  Plu, Canwen Xu, Teven Le~Scao, Sylvain Gugger, Mariama Drame, Quentin Lhoest,
  and Alexander Rush. 2020.
\newblock \href {https://doi.org/10.18653/v1/2020.emnlp-demos.6} {Transformers:
  State-of-the-art natural language processing}.
\newblock In \emph{Proceedings of the 2020 Conference on Empirical Methods in
  Natural Language Processing: System Demonstrations}, pages 38--45.
  Association for Computational Linguistics.

\bibitem[{{World Health Organization}(2022)}]{WHOreport}
{World Health Organization}. 2022.
\newblock \href
  {https://www.who.int/publications/m/item/weekly-epidemiological-update-on-covid-19---16-november-2022}
  {Weekly epidemiological update on {COVID}-19 - 16 {November} 2022}.

\bibitem[{Wouters et~al.(2020)Wouters, McKee, and Luyten}]{WoutersDrugDev}
Olivier~J. Wouters, Martin McKee, and Jeroen Luyten. 2020.
\newblock \href {https://doi.org/10.1001/jama.2020.1166} {Estimated research
  and development investment needed to bring a new medicine to market,
  2009-2018}.
\newblock \emph{JAMA}, 323(9):844--853.

\bibitem[{Xu et~al.(2020)Xu, Zhang, Mao, Wang, Xie, and Zhang}]{nlucurriculum}
Benfeng Xu, Licheng Zhang, Zhendong Mao, Quan Wang, Hongtao Xie, and Yongdong
  Zhang. 2020.
\newblock \href {https://doi.org/10.18653/v1/2020.acl-main.542} {Curriculum
  learning for natural language understanding}.
\newblock In \emph{Proceedings of the 58th Annual Meeting of the Association
  for Computational Linguistics}, pages 6095--6104. Association for
  Computational Linguistics.

\bibitem[{Yu et~al.(2020)Yu, Li, Chen, Zhou, Wang, Li, Jiang, and
  Wang}]{hcqPositive}
Bo~Yu, Chenze Li, Peng Chen, Ning Zhou, Luyun Wang, Jia Li, Hualiang Jiang, and
  Dao~Wen Wang. 2020.
\newblock \href {https://doi.org/10.1101/2020.04.27.20073379}
  {Hydroxychloroquine application is associated with a decreased mortality in
  critically ill patients with {COVID}-19}.
\newblock \emph{medRxiv}.

\end{thebibliography}
\bibliographystyle{acl_natbib}

\newpage
\clearpage
\onecolumn
\appendix

\section*{Supplementary Materials}

\section{Further Details about the COVID-19 NLI Dataset}

In this appendix we provide additional details about the creation of the COVID-19 NLI dataset. Our annotators are experts in the domain having trained as healthcare providers (nursing) and annotation. The research annotator (DS) is a specialist in the biomedical domain with background in molecular biology and computer science. Annotators have also provided span annotations in several cases of drug mention, polarity, context, and expressions of uncertainty to aid in the annotation task.

\subsection{Inter-Annotator Analysis}
\label{sec:iaaAndGuidelines}

Two rounds of inter-annotator analysis were conducted to converge on a set of annotation guidelines for scaling and to measure consistency between multiple annotators. In the first round four annotators (three clinical annotators, one researcher) were presented with 64 pairs of extracted research claims and an initial set of annotation guidelines. Classification was conducted across five classes including a Strict Entailment and Strict Contradiction class indicating two claims were entailing or contradicting in a strict logical sense as opposed to a common-reasoning sense. Global Fleiss' $\kappa$ for this round was 0.83. For the second round, three annotators (two clinical annotators, one researcher) annotated 75 claim pairs with updated guidelines and achieved similar consistency at $\kappa = 0.84$.
Further minor modifications were made to the annotation guidelines resulting in the final guidelines used for the scaling round (Table \ref{tab:Tab1}).

\newcolumntype{L}{>{\arraybackslash}m{6cm}}
\definecolor{Con}{HTML}{D81B74}
\definecolor{Ent}{HTML}{52CC62}
\definecolor{Neu}{HTML}{949494}

\begin{table*}[hp]
\centering
\setlength{\tabcolsep}{12pt}
\begin{tabular}{L c}
\toprule
\textbf{Criteria}                                                                            & \textbf{Annotation}                   \\ \midrule
All drugs, context, and   sentiment match                                                    & \textcolor{Ent}{STRICT ENTAILMENT}    \\ \hline
At least one drug matches, the   sentiment is the same but the context is at least   similar & \textcolor{Ent}{ENTAILMENT}                            \\ \hline
All drugs and context match but   the sentiment is opposing                                  & \textcolor{Con}{STRICT CONTRADICTION}                  \\ \hline
At least one drug matches, the   sentiment is opposing but the context is at least similar   & \textcolor{Con}{CONTRADICTION}                         \\ \hline
The context or sentiment   statement cannot be compared                                      & \textcolor{Neu}{NEUTRAL}                              \\ \hline
There is no mention of a drug   OR none of the drugs match                                   & \textcolor{Neu}{NEUTRAL}                               \\ \hline
One claim contains both a POSITIVE and a NEGATIVE statement and the other claim contains a POSITIVE or NEGATIVE statement                                & \textcolor{Con}{CONTRADICTION}                            \\ \hline
One claim is POSITIVE or   NEGATIVE statement and the other is EXPRESSION\_OF\_UNCERTAINTY   & \textcolor{Neu}{NEUTRAL}                               \\ \hline
Both claims are EXPRESSION\_OF\_UNCERTAINTY                                                  & \textcolor{Ent}{ENTAILMENT}                \\ \bottomrule
\end{tabular}
\caption{Annotation guidelines for the COVID-19 NLI dataset.}
\label{tab:Tab1}
\end{table*}

\subsection{Preparing Claims for Annotation}
\label{sec:claimPrep}

We used a set of heuristics to generate a set of non-trivial (obviously neutral) claim pairs for annotators to label. First, we considered three topics, $t \in T =$ \{ \text{``mortality''}, \text{``effective treatment''}, \text{``toxicity''}\} and seven drugs, $d \in D =$ \{\text{``hydroxychloroquine''}, \text{``chloroquine''}, \text{``tocilizumab''}, \text{``remdesivir''}, \text{``vitamin D''}, \text{``lopinavir''}, \text{``dexamethasone''}\}. For each pair, $(t, d)$, the following procedure (Algorithm \ref{alg:pairs}) was used to generate candidate claim pairs from the set of true research claims, $C$. Additionally given \textit{pol}(.), a function for calculating the polarity of a claim; $k$, the number of claims to sample that are relevant to a drug and topic and have a given polarity (positive or negative); and $N$, the total number of pairs to subsample, we define our heuristic algorithm for generating candidate non-trivial pairs in Algorithm \ref{alg:pairs}.

\renewcommand{\algorithmicrequire}{\textbf{Input:}}
\renewcommand{\algorithmicensure}{\textbf{Output:}}

\begin{algorithm}[H]
\caption{Heuristic sampler for generating candidate non-trivial pairs}
\label{alg:pairs}
\begin{algorithmic}[1]
\Require{Topic set $T$, drug set $D$, claim set $C$, polarity function $\textit{pol}(.) : c \rightarrow [-1,1]$, drug topic claim sample size $k$, total subsample size $N$}
\Ensure{Set of $N$ claim pairs $P_N$ concerning a common drug and topic and non-neutral predicted polarity}
\State $P \gets \emptyset$
\For{$(d,t) \in D \times T$}
    \State {Retrieve claims $C_d := \{c \in C : d \text{ is a substring of } c\}$}
    \State {Define $C_{d,t,k,\textit{pos}}$ := top $k$ claims $c$ relevant to $t$ from $C_d$ s.t.\ $\textit{pol}(c) > 0$}
    \State {Define $C_{d,t,k,\textit{neg}}$ := top $k$ claims $c$ relevant to $t$ from $C_d$ s.t.\ $\textit{pol}(c) < 0$}
    \State {Enumerate all combinations of claim pairs, $P_{d,t,2k}$, from claims in set  $C_{d,t,k,\textit{pos}} \cup C_{d,t,k,\textit{neg}}$}
    \State {Remove copy claim pairs, $P_{d,t,2k} \gets P_{d,t,2k} \setminus \{(c_1, c_2) \in P_{d,t,2k} : c_1 = c_2\}$}
    \State {$P \gets P \cup P_{d,t,2k}$}
\EndFor
\State {Sample $N$ pairs uniformly from $P$, $P_N$}
\end{algorithmic}
\end{algorithm}

We set $k=7$ and $N=1000$. To evaluate claim relevancy (lines 4 and 5), we calculate the cosine similarity between an embedding of the topic and sentence embeddings of claims using uSIF \cite{usif}. Polarity, $\textit{pol}(.)$, is calculated using Vader scores \cite{vader}.

\section{Curriculum Datasets}
\label{sec:curriculumDatasets}
We included four datasets for fine-tuning our language models, which comprise general language and multiple biomedically-focused domains. All our datasets use the labels Entailment, Contradiction, and Neural. For our COVID-19 NLI dataset, we collapse Strict Entailment with Entailment and Strict Contradiction with Contradiction. \par

\subsection{MultiNLI}
MultiNLI is an NLI dataset consisting of 433k premise-hypothesis pairs taken from 5 general domains \citep{multinli}. To create the dataset, annotators were shown a premise and were asked to provide hypothesis statements that were entailed by, contradicted by, or were neutral to the prompt premise. In this work, we used the \textit{matched} validation set for evaluation, which we split into two equal sized validation and test sets.

\subsection{MedNLI}
MedNLI is an NLI dataset consisting of 14k premise-hypothesis pairs where premises are extracted from doctor's notes in electronic medical records \citep{mednli}. The annotation task for generating premise-hypothesis pairs was analogous to that for MultiNLI.

\subsection{ManConCorpus}
ManConCorpus is a dataset of research claims taken from biomedical systematic reviews \citep{manConCorpus}. These reviews compile together studies that investigate a common research questions and consider their findings in aggregate. The research question, which conforms to the standardized PICO criteria \citep{picoquestion}, yields a binary answer, so findings from the associated review will take explicit ``yes'' or ``no'' stances. One such PICO question is ``In elderly populations, does omega 3 acid from fatty fish intake, compared with no consumption, reduce the risk of developing heart failure?'' \citep{LiFishConsumption}. \par
Pairs of claims manually annotated from these works can be paired together for NLI classification by matching claims that take the same stance on a common question as entailing pairs, those that take opposite stances on a common question as contradicting pairs, and those taken from two different reviews about different questions as neutral pairs. The dataset's 16 PICO questions are split into 12, 4, and 4 questions for the train, validation, and test splits, respectively, and the neutral class is downsampled to be the same size as the next largest class in all splits. The resulting dataset has 2.8k claim pairs in total.

\section{Curriculum Design}
\label{sec:curriculumDesign}
To create an effective curriculum for the ultimate task of detecting contradictions in the COVID-19 treatment domain, we conducted a set of experiments analyzing the effect of multiple design decisions for incorporating domain-adjacent corpora in training.

\subsection{Experiments}

\subsubsection{Shuffled and Combined Curricula}
\label{sec:shuffComb}
To understand the importance of sequencing the curriculum, we evaluated BERT models trained using various sequences of domain-adjacent corpora in equal proportion. We consider three types of curricula: forward, reverse, and shuffled. The forward curriculum proceeds with fine-tuning a pre-trained BERT model in sequence from the most general domain (MultiNLI) to MedNLI to ManConCorpus to the most relevant domain (COVID-19 NLI). The reverse curriculum begins with the most relevant domain and proceeds in the opposite direction. The shuffled curricula were sampled from the 22 possible random orderings of the four domains excluding the forward and reverse sequences. We sampled three shuffled domains to assess the background from non-intentional curriculum design. Finally, we considered a ``combined'' curriculum where data from the four corpora are concatenated together and shuffled, thus ablating the notion of intentional sequencing in the curriculum. To ensure no dataset dominated training, each dataset, $D_{\textit{train}}$, is subsampled such that $N_{D_{\textit{train}}} = \min{(d, |D_{\textit{train}}|)}$ samples are present in the curriculum.

\subsubsection{Ordered Curriculum Subsequence Fine-Tuning}
\label{sec:orderedForward}
To assess the contribution to performance from specific domains during sequencing as well as the effect of curriculum size, we evaluated forward curriculum subsequences. Ten subsequences were evaluated: the full forward curriculum, two three-dataset subsequences, three two-dataset subsequences, and the four single corpora. As in \ref{sec:shuffComb}, $N_{D_{\textit{train}}}$ samples are present in the curriculum from dataset $D_{\textit{train}}$.

\subsubsection{Perturbing Dataset Proportion in Sequential Curricula}
\label{sec:dataRatio}
To assess whether changing the ratio of training data used from the various corpora yielded better performance or to dilutive biases from larger corpora, we modulated the data ratio parameter. We define data ratio, $r$, as the multiplicative factor larger a dataset is from the next dataset in the curriculum sequence. Specifically, given $r$, we calculate the sample size of the dataset, $D_{\textit{train}}$, to be used in the $i$th step (1-index) of a size-$k$ fine-tuning curriculum as $N_{D_{\textit{train}}} = \min{(r^{k-i}d, |D_{\textit{train}}|)}$. We considered three curricula: the full forward curriculum and the two sequential three-dataset curricula.

\subsection{Evaluation}

\begin{figure}
\centering
\includegraphics[width=.75\textwidth]{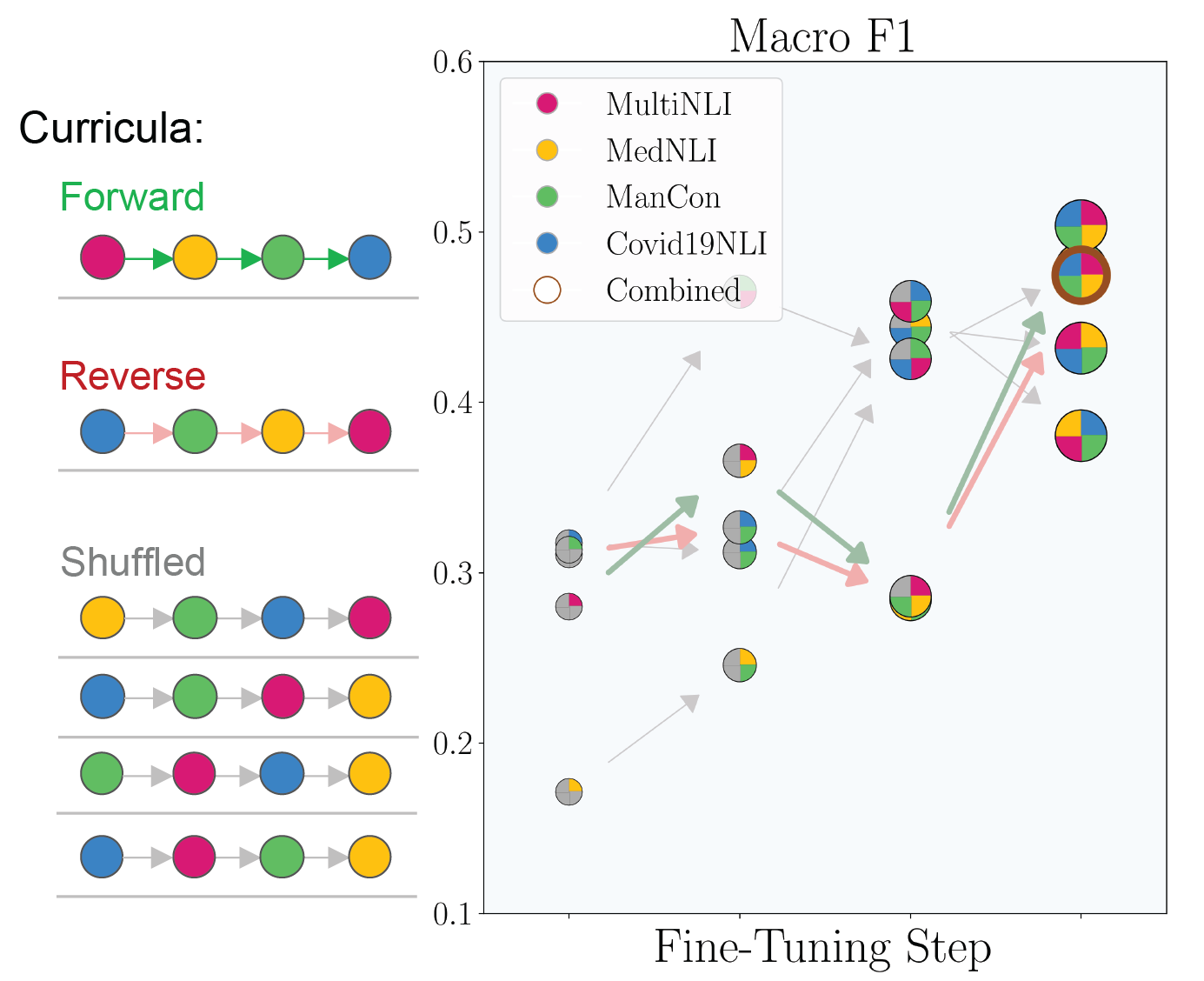}
\caption{Macro F1 evaluation on the COVID-19 NLI validation set for different orderings of the fine-tuning curriculum and the combined curriculum (brown). Pie point position indicate F1 value, colors indicate which corpora have been already introduced to the model at a given fine-tuning step, and arrows indicate curriculum sequences. Pie size is proportional to amount of training data seen by the model.}
\label{fig:longSeqs}
\end{figure}

\subsubsection{Setup}
We evaluated multiple BERT models pre-trained using general and biomedical corpora for curriculum-based fine-tuning \cite{bert}. Each fine-tuning step involves training for 4 epochs with learning rate $l = 10^{-5}$ and batch size $b = 8$. For all experiments, $d = 500$, and for data ratio experiments, $r\in \{1, 2\}$. Pre-trained models were loaded from the HuggingFace Transformers library \cite{huggingface}. All fine-tuning was conducted on 2 Tesla-V100-SXM2 and 2 Tesla-A100-PCIe GPUs. Experiments in curriculum design were evaluated with the pre-trained PubMedBERT model \citep{pubmedbert}. Other pre-trained BERT models were evaluated on forward curriculum subsequences (Appendix \ref{sec:bertPretraining}).

\subsubsection{Evaluation Metrics}
The primary NLI evaluation metric for fine-tuned BERT models was macro F1 on the COVID-19 NLI validation set. We also investigated recall of the contradictions class as an important metric in evaluating the ability to detect contradictory research claims.

\subsubsection{Shuffled and Collapsed Curricula}
Of the six tested four-dataset curricula, the forward curriculum performed highest with an F1 of 0.503. The reverse curriculum, starting with the most relevant and challenging curriculum first, achieved an F1 of 0.474. The shuffled curricula yielded F1 scores of 0.380, 0.432, and 0.478. The collapsed curriculum, in which the four corpora are concatenated and shuffled, achieved competitive performance as well, yielding an F1 score of 0.475 (Figure \ref{fig:longSeqs}).

\subsubsection{Ordered Subsequences}

\begin{figure}
\centering
\includegraphics[width=.75\textwidth]{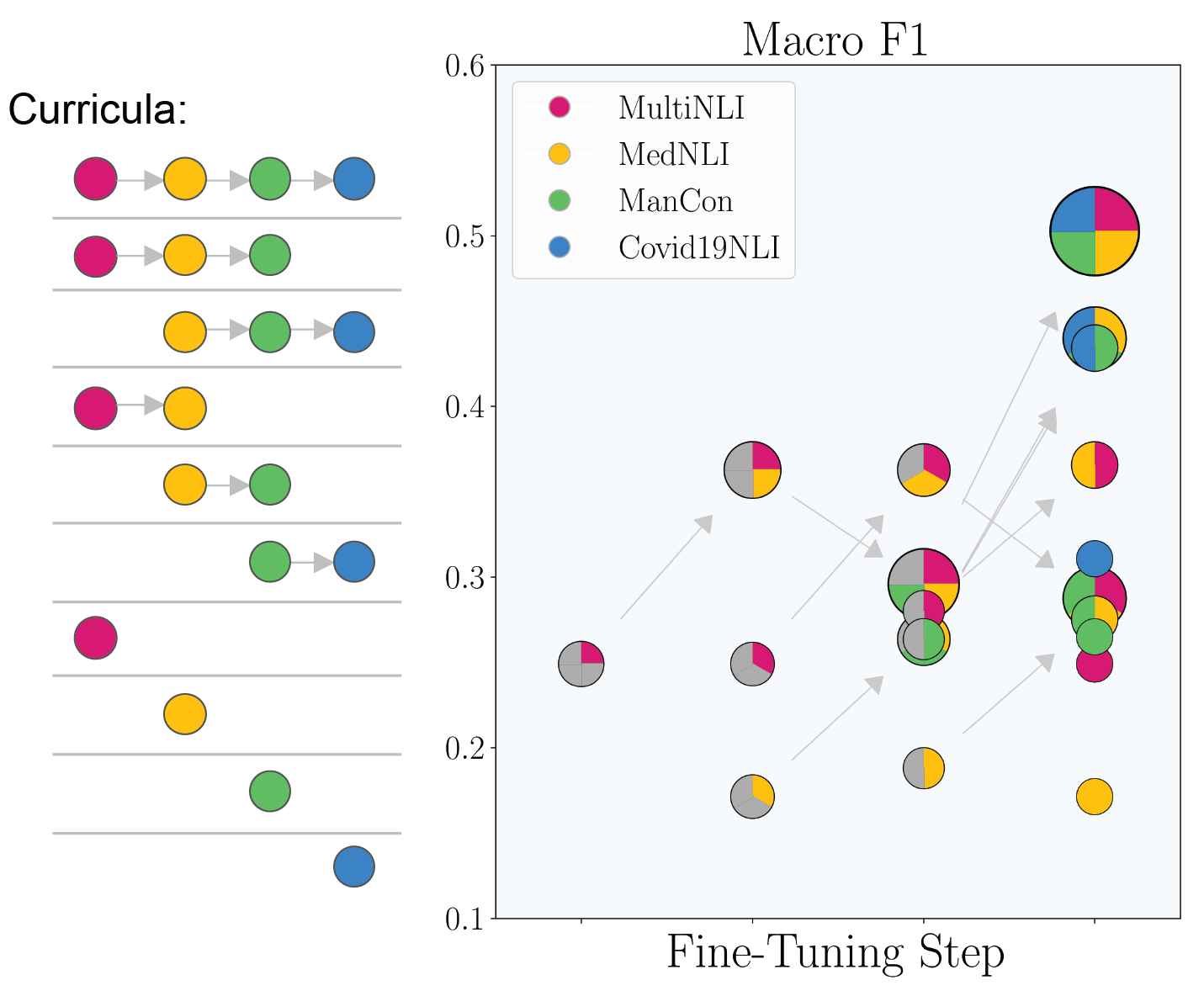}
\caption{Macro F1 evaluation for various subsequences of the forward curriculum.}
\label{fig:forwardSubseqs}
\end{figure}

From the 10 curriculum subsequences, the model trained with the full forward curriculum yielded highest performance with an F1 of 0.503. Among the two three-domain sequences, the one including the in-domain COVID-19 NLI dataset achieved greater performance than that without, yielding F1 scores of 0.440 and 0.296 respectively. Similarly, with the two-domain subsequences, the sequence with ManConCorpus and COVID-19 performed best with F1 of 0.434, and the subsequence containing MedNLI and ManConCorpus performed worst with F1 of 0.275. Among the single domain curricula, the in-domain training on our dataset was best with F1 of 0.311 (Figure~\ref{fig:forwardSubseqs}).

\subsubsection{Variable Dataset Proportions}

\begin{figure}
\centering
\includegraphics[width=.75\textwidth]{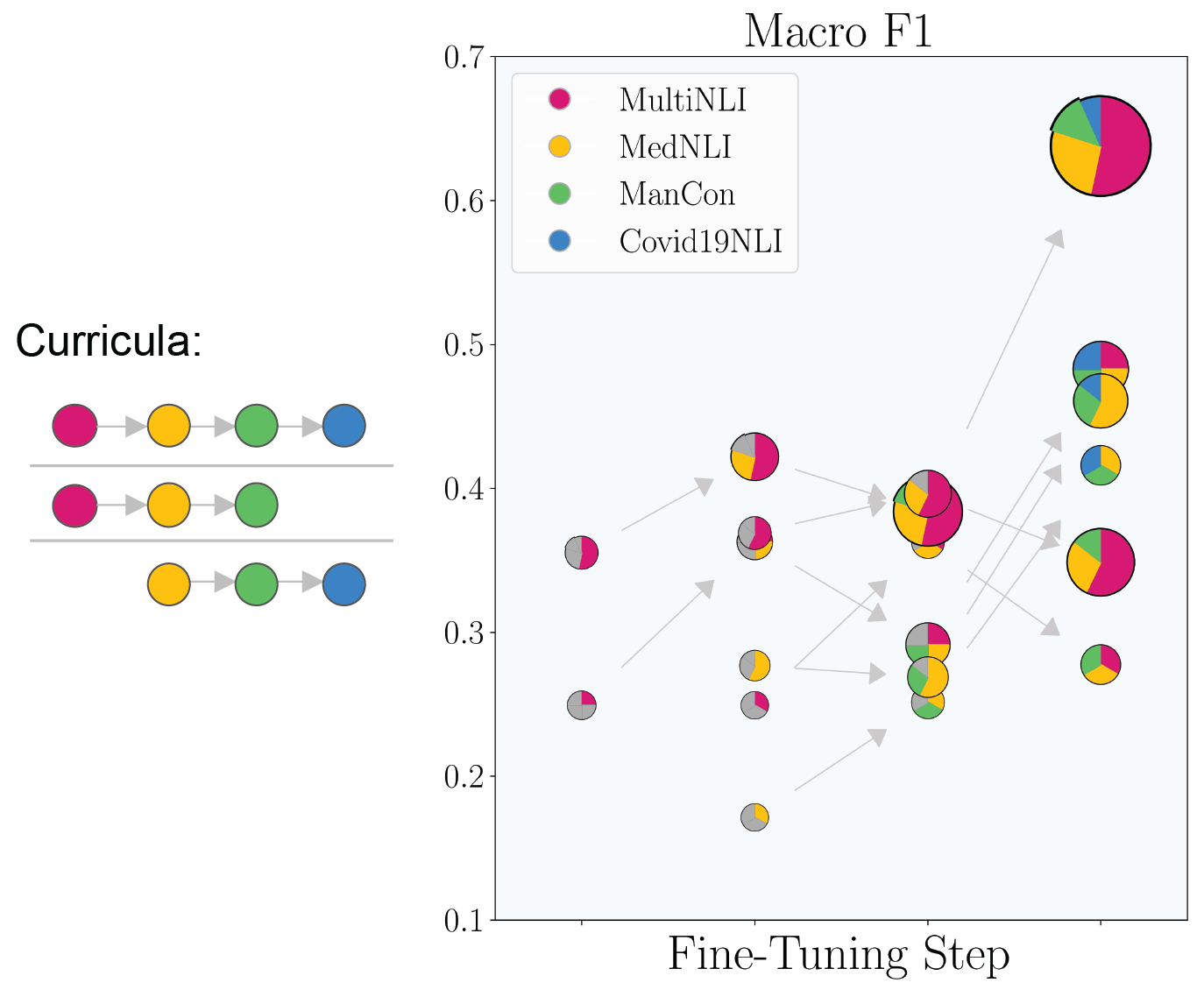}
\caption{Evaluation of three subsequences of the forward curriculum with two different data ratio proportions, $r \in \{1, 2\}$.}
\label{fig:dataRatios}
\end{figure}

In all three curricula, the condition with data ratio $r = 2$ outperformed the $r = 1$ equal data proportion condition. The highest performing curriculum was the $r = 2$ forward curriculum achieving an F1 of 0.638. In the in-domain three-dataset sequence, F1 increased from 0.416 with $r=1$ to 0.461 with $r=2$. The out-of-domain three-sequence saw a similar increase in performance favoring $r=2$ over $r=1$ (Figure~\ref{fig:dataRatios}).

\section{BERT Pretraining}
\label{sec:bertPretraining}

\begin{figure*}
\includegraphics[width=\textwidth]{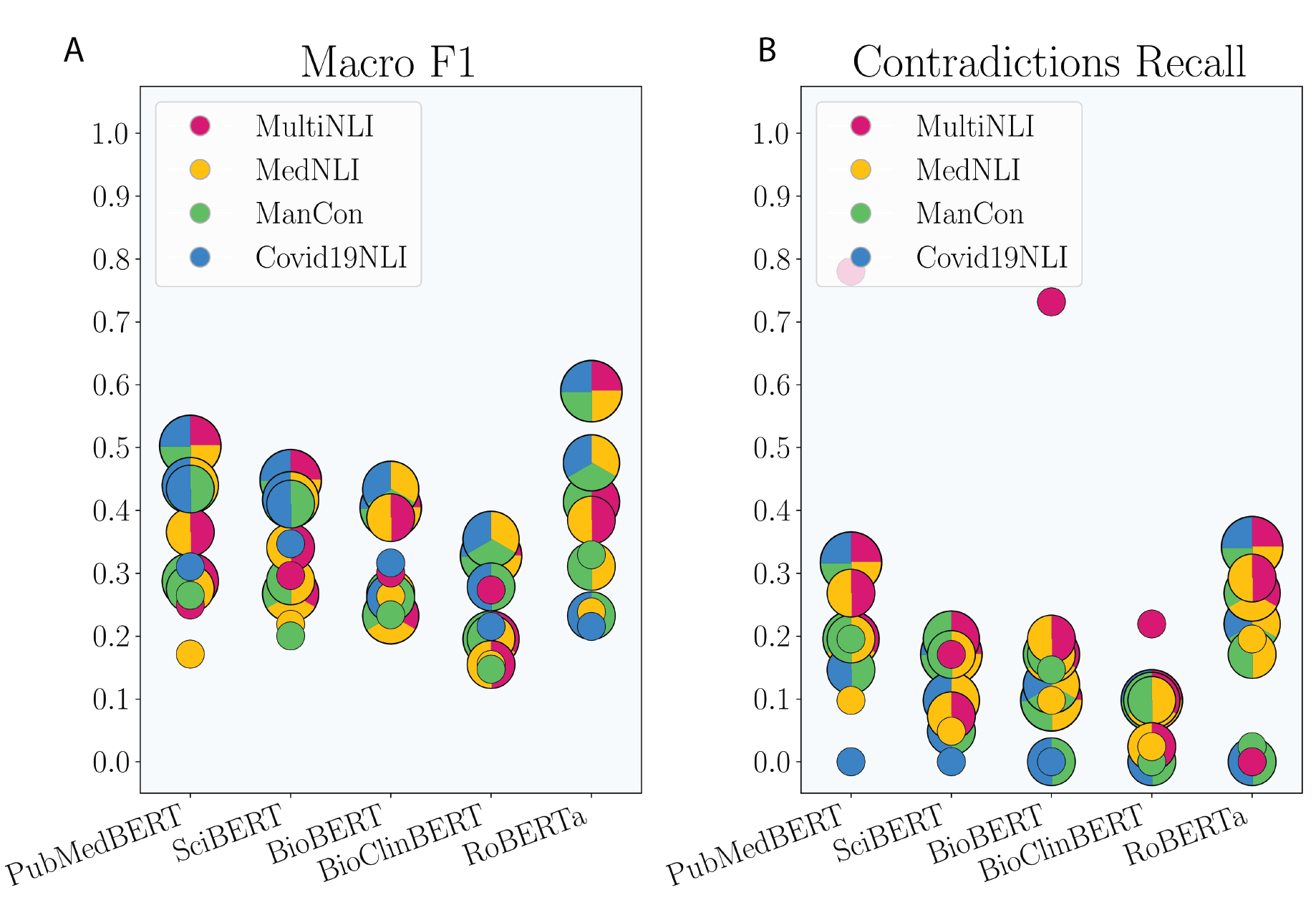}
\caption{Macro F1 (A) and contradictions recall (B) for five pre-trained BERT models: PubMedBERT, SciBERT, BioBERT, BioClinBERT, and RoBERTA fine-tuned with subsequences of the forward curriculum.}
\label{fig:modelComps}
\centering
\end{figure*}

Five pre-trained BERT models were evaluated for further fine-tuning: PubMedBERT \citep{pubmedbert}, SciBERT \citep{scibert}, BioBERT \citep{biobert}, BioClinBERT \citep{bioclinbert}, and RoBERTA \cite{roberta}. We conducted fine-tuning experiments under the same 10 subsequences and parameter settings as in Section \ref{sec:orderedForward} and evaluated performance on the validation split of the COVID-19 NLI dataset. For PubMedBERT, SciBERT, and RoBERTa, the full forward curriculum yielded the greatest macro F1 scores at 0.503, 0.448, and 0.590, respectively. The greatest performance was achieved by the MedNLI-ManCon-COVID-19 NLI subsequence for BioBERT and BioClinBERT models yielding F1 scores of 0.433 and 0.354 (Figure \ref{fig:modelComps}).

\section{BERT Hyperparameter Tuning}
\label{sec:hpOpt}

\begin{figure*}
\centering
\includegraphics[width=\textwidth]{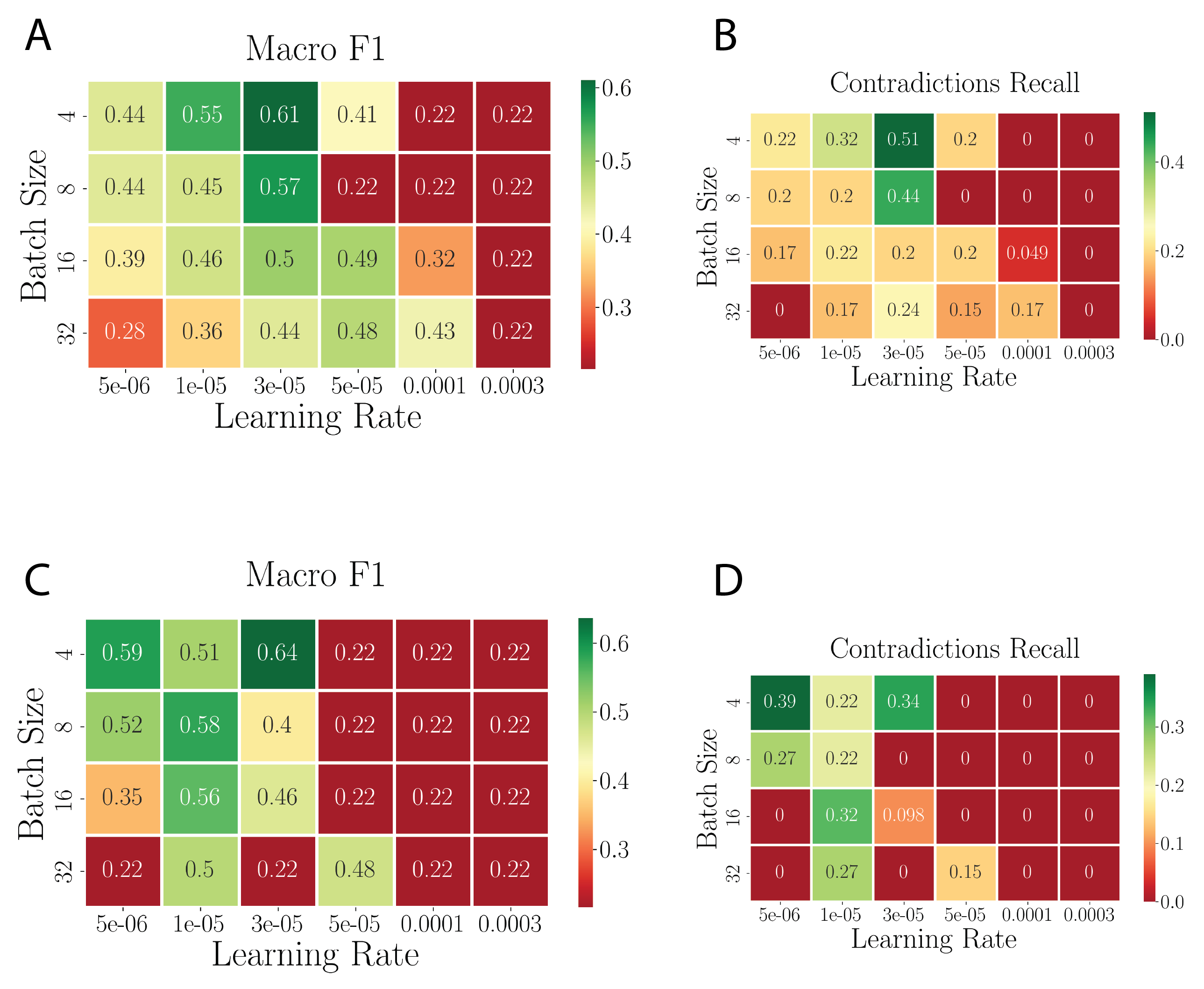}
\caption{Batch size and learning rate hyperparameter optimization evaluated with the COVID-19 NLI validation metrics for PubMedBERT (A, B) and RoBERTa (C, D) models.}
\label{fig:HPTune}
\end{figure*}

We evaluated macro F1 and contradictions recall on the COVID-19 NLI validation set over a parameter sweep of learning rates, $lr \in \{\num{5e-6}, \num{1e-5}, \num{3e-5}, \num{5e-5}, \num{1e-4}, \num{3e-4}\}$ and batch sizes, $b \in \{4, 8, 16, 32\}$ for PubMedBERT and RoBERTa models. For both models the highest macro F1 setting was $lr = \num{3e-5}$ and $b = 4$ yielding $F1 = 0.61$ and $F1 = 0.64$ for PubMedBERT and RoBERTa, respectively. These settings yielded the greatest contradictions recall of 0.51 for PubMedBERT, and settings of $lr = \num{5e-6}, b = 4$ yielded the highest contradictions recall value of 0.39 for RoBERTa (Figure \ref{fig:HPTune}).

\section{Test Set Evaluation and Baselines}\label{sec:testSetEval}

We evaluated test set statistics for the COVID-19 NLI using PubMedBERT and RoBERTa \cite{roberta} models fine-tuned with the forward curriculum of MultiNLI $\rightarrow$ MedNLI $\rightarrow$ ManCon $\rightarrow$ COVID-19 NLI. We set data ratio as being equal between the four corpora ($r=1$) (see Appendix \ref{sec:dataRatio}), and after hyperparameter tuning of learning rate and batch size (Appendix \ref{sec:hpOpt}) set parameters $l_{\textit{HP}} = 3 * 10^{-5}$ and $b_{\textit{HP}} = 4$.

We compared performance of our trained BERT models to several NLI baselines.
\begin{itemize}
  \item {\textbf{Hypothesis-Only Unigrams} Softmax classification using unigram counts in the hypothesis (single claim).}
  \item {\textbf{Word Overlap} Softmax classification over counts of overlapping unigrams from the two claims.}
  \item {\textbf{Word Cross-Product} Softmax classification over counts of pairs of words in the cross-product between the two claims.}
  \item {\textbf{Similarity + Polarity} Softmax classification using similarity of the two claims as calculated using uSIF sentence embeddings \cite{usif, fse} and polarity of each claim using Vader polarity scores \cite{vader}.}
  \item {\textbf{Hypothesis-Only BERT} BERT classification where one of the two claims has been ablated.}
\end{itemize}

\begin{figure*}[hp]
\centering
\includegraphics[width=\textwidth]{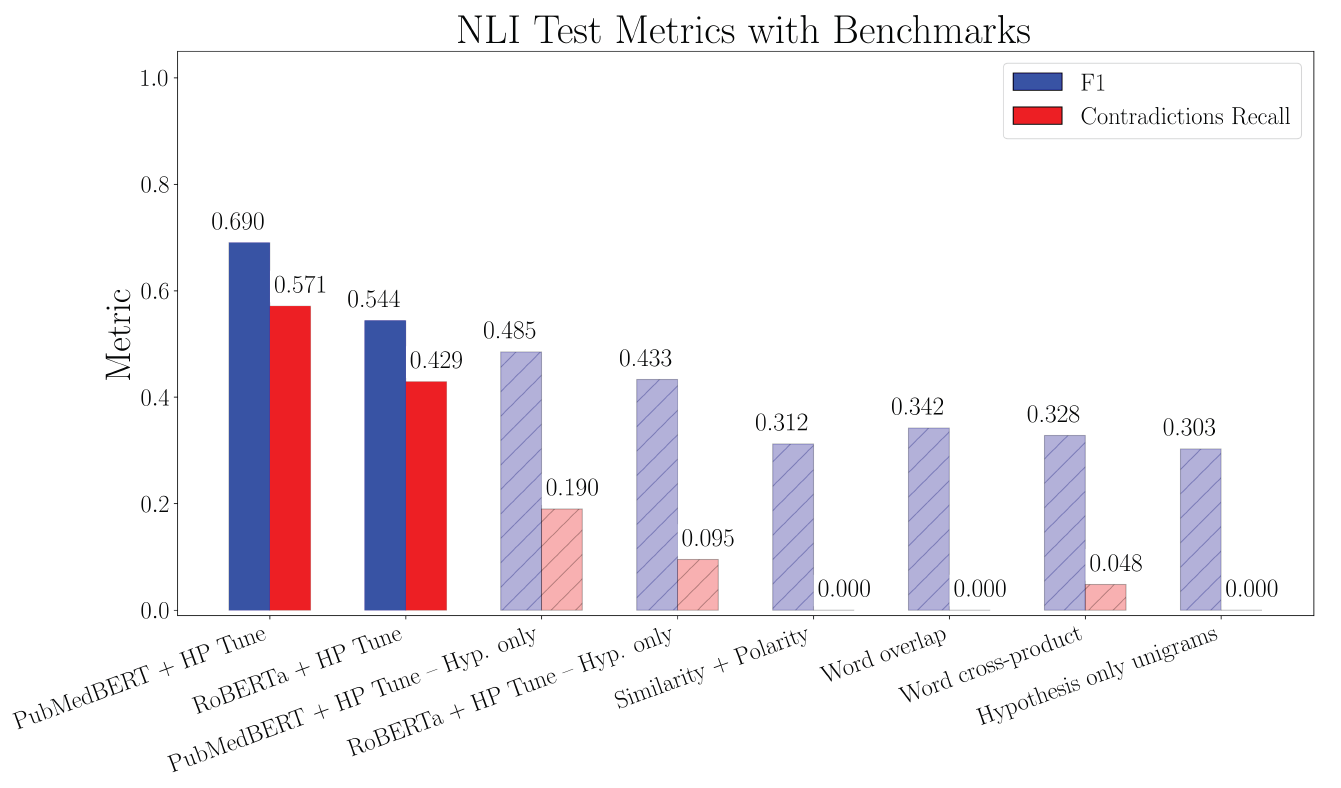}
\caption{Macro F1 and contradictions recall for PubMedBERT and RoBERTa models fine-tuned with the forward curriculum compared to NLI benchmarks. `HP Tune' indicates hyperparameter tuning (Appendix~\ref{sec:hpOpt}).}
\label{fig:testMetrics}
\end{figure*}

Figure~\ref{fig:testMetrics} offers a comparison of these baselines with our proposed models, focusing on the forward curriculum condition. We also evaluated the optimized PubMedBERT and RoBERTa models with the reverse curriculum and four shuffled curricula \ref{tab:RevShuffBERTs}. We note the consistent result that the forward curriculum performs best overall.

\begin{table}[tp]
\centering
\resizebox{0.75\columnwidth}{!}{
\setlength{\tabcolsep}{4pt}
\begin{tabular}{@{} l l r r @{}}
\toprule
 & & & Contra. \\
Model & Curriculum & F1 &  Recall \\
\midrule
\multirow{6}{*}{PubMedBERT} & \text{Multi} $\rightarrow$ \text{Med} $\rightarrow$ \text{ManCon} $\rightarrow$ \text{Covid} & \textbf{0.690} & \textbf{0.571} \\
& \text{Covid} $\rightarrow$ \text{ManCon} $\rightarrow$ \text{Med} $\rightarrow$ \text{Multi} & 0.428 & 0.381 \\
& \text{Covid} $\rightarrow$ \text{Multi} $\rightarrow$ \text{ManCon} $\rightarrow$ \text{Med} & 0.486 & 0.381 \\
& \text{Covid} $\rightarrow$ \text{ManCon} $\rightarrow$ \text{Multi} $\rightarrow$ \text{Med} & 0.581 & 0.571 \\
& \text{Med} $\rightarrow$ \text{ManCon} $\rightarrow$ \text{Covid} $\rightarrow$ \text{Multi} & 0.446 & 0.381 \\
& \text{ManCon} $\rightarrow$ \text{Multi} $\rightarrow$ \text{Covid} $\rightarrow$ \text{Med} & 0.579 & 0.333 \\
\midrule
\multirow{6}{*}{RoBERTa} & \text{Multi} $\rightarrow$ \text{Med} $\rightarrow$ \text{ManCon} $\rightarrow$ \text{Covid} & 0.544 & 0.429 \\
 & \text{Covid} $\rightarrow$ \text{ManCon} $\rightarrow$ \text{Med} $\rightarrow$ \text{Multi} & 0.411 & 0.476 \\
& \text{Covid} $\rightarrow$ \text{Multi} $\rightarrow$ \text{ManCon} $\rightarrow$ \text{Med} & 0.319 & 0.476 \\
& \text{Covid} $\rightarrow$ \text{ManCon} $\rightarrow$ \text{Multi} $\rightarrow$ \text{Med} & 0.232 & 0 \\
& \text{Med} $\rightarrow$ \text{ManCon} $\rightarrow$ \text{Covid} $\rightarrow$ \text{Multi} & 0.174 & 0 \\
& \text{ManCon} $\rightarrow$ \text{Multi} $\rightarrow$ \text{Covid} $\rightarrow$ \text{Med} & 0.232 & 0 \\
\bottomrule
\end{tabular}
}
\caption{Test set performance on optimized PubMedBERT and RoBERTa models trained with various fine-tuning curricula.}
\label{tab:RevShuffBERTs}
\end{table}

\end{document}